\documentclass{article}
\usepackage{spconf,amsmath,graphicx,hyperref}
\usepackage{multirow, booktabs}
\usepackage{amsfonts, enumitem, amssymb}
\usepackage{algorithm, algpseudocode}

\usepackage[table]{xcolor}
\usepackage{xcolor}
\usepackage{duckuments}
\usepackage{csquotes}

\newcommand{\gr}{\rowcolor{gray!10}}
\definecolor{BrickGreen}{rgb}{0,0.6,0}
\definecolor{BrickGray}{rgb}{0.5,0.5,0.5}
\definecolor{BrickRed}{rgb}{0.6,0,0}

\title{On the Importance of a Multi-Scale Calibration for Quantization}

\name{Seungwoo Son, Ingyu Seong, Junhan Kim, Hyemi Jang and Yongkweon Jeon$^{\dagger}$\thanks{$^{\dagger}$Corresponding Author}}
\address{Samsung Research, Seoul, South Korea
\\ \normalsize{swson32@gmail.com, \{ingyu.seong, jun\_one.kim, hye\_mi.jang, dragwon.jeon\}@samsung.com}}

\begin{document}

\maketitle
\begin{abstract}

Post-training quantization (PTQ) is a cornerstone for efficiently deploying large language models (LLMs), where a small calibration set critically affects quantization performance. However, conventional practices rely on random sequences of fixed length, overlooking the variable-length nature of LLM inputs. Input length directly influences the activation distribution and, consequently, the weight importance captured by the Hessian, which in turn affects quantization outcomes. As a result, Hessian estimates derived from fixed-length calibration may fail to represent the true importance of weights across diverse input scenarios.
We propose MaCa (\underline{Ma}tryoshka \underline{Ca}libration), a simple yet effective method for length-aware Hessian construction. MaCa (i) incorporates multi-scale sequence length information into Hessian estimation and (ii) regularizes each sequence as an independent sample, yielding a more stable and fruitful Hessian for accurate quantization.
Experiments on state-of-the-art LLMs (e.g., Qwen3, Gemma3, LLaMA3) demonstrate that MaCa consistently improves accuracy under low bit quantization, offering a lightweight enhancement compatible with existing PTQ frameworks. To the best of our knowledge, this is the first work to systematically highlight the role of multi-scale calibration in LLM quantization.
\end{abstract}

\section{Introduction}
While large language models (LLMs) deliver remarkable capabilities, they also impose substantial computational demands. Leading LLMs typically contain hundreds of billions of parameters, requiring extensive memory for inference. For instance, gpt-oss-120b \cite{agarwal2025gpt}, one of the most recent high performing models, consumes no less than 80 GB of memory.

\begin{figure}[t]
    \label{fig:main-figure}
    \centering
    \includegraphics[width=0.49\textwidth]{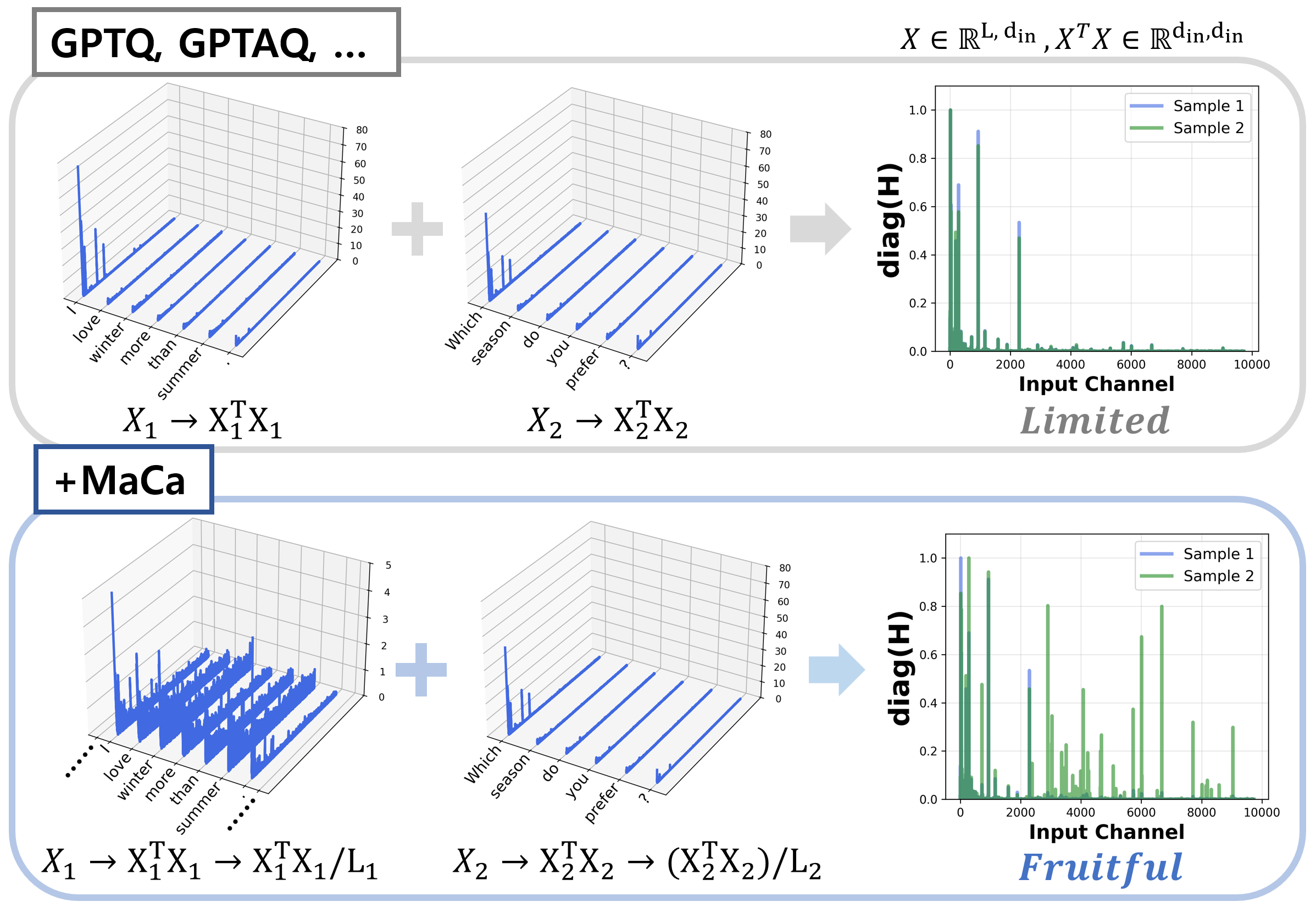}
    \vspace{-2em}
    \caption{\textbf{Visualization of diagonals of Hessian.} Top: GPTQ with fixed length sequences makes limited Hessian diagonals. Bottom: MaCa with varied lengths produces a richer Hessian that captures diverse channel sensitivities.}
\vspace{-1em}
\end{figure}

To alleviate such costs, recent research has investigated quantization techniques \cite{son2024prefixing} that reduce weight or activation precision while preserving full precision accuracy. One line of work employs gradient to minimize the quantization error \cite{nagel2020adaround, li2021brecq}. These methods, however, are too time-consuming, which severely limits their scalability, especially for LLMs with billions of parameters. A more practical approach, spearheaded by GPTQ \cite{frantar2023gptq} and followed by subsequent works \cite{kim2025boa}, has gained broad adoption by approximating Hessian information. The Hessian can be expressed as $H_{(w)}=(X X^\top) \otimes \nabla_{z}^{2}\mathcal{L} = H_{\mathrm{in}} \otimes H_{\mathrm{out}}$ where $X$ denotes the input activation, $\nabla_{z}^{2}\mathcal{L}$ is the second derivative of the loss with respect to the pre-activation $z$. GPTQ and its extensions \cite{kim2025boa, kim2024towards} focus on better approximating $H_{out}$. 

However, they overlook a key aspect of $H_{in}$. Typically, $H_{in} = XX^T$ with $X \in \mathbb{R}^{D \times L}$ and activations vary greatly with sequence length. As shown in Figure~\ref{fig:main-figure}, LLM activation distributions strongly depend on input length, and the Hessian varies significantly with sequence length. This limitation is further illustrated in the 1D plot in Figure~\ref{fig:main-figure}, which shows the Hessian diagonals of the \texttt{down-proj} in the final MLP block of Qwen3-4B. Specifically, longer sequences mainly emphasize early input channels, whereas shorter sequences also focus on later channels. These complementary patterns remain invisible under fixed length calibration. 
%
This naturally leads to the key question:

\begin{displayquote}
\textit{Can we design a calibration strategy that makes the Hessian robust to varying sequence lengths?}
\vspace{-1.0em}
\end{displayquote}

\begin{table*}[ht]
\caption{\textbf{Average accuracy (\%)} over 8 zero-shot benchmark tasks for GPTQ and GPTAQ, with and without MaCa, at 4bit, 3bit, and 2bit/g128 across Qwen3, Gemma3, and LLaMA3. Results are reported as mean ± std over 3 seeds.}
\small
\centering
\resizebox{0.98\textwidth}{!}{%
\begin{tabular}{l|ccc|ccc|ccc}
\toprule
\textbf{Method} & \multicolumn{3}{c|}{\textbf{Qwen3-4B-IT}} & \multicolumn{3}{c|}{\textbf{Gemma3-4B-IT}} & \multicolumn{3}{c}{\textbf{LLaMA3.2-3B-IT}} \\ \midrule
& 4bit & 3bit & 2bit/g128 & 4bit & 3bit & 2bit/g128 & 4bit & 3bit & 2bit/g128 \\ \midrule
GPTQ   & 55.01{\scriptsize$\pm$0.60} & 41.99{\scriptsize$\pm$0.27} & 43.23{\scriptsize$\pm$0.64} & 
          50.94{\scriptsize$\pm$0.46} & 43.38{\scriptsize$\pm$0.62} & 38.66{\scriptsize$\pm$0.14} & 
          33.20{\scriptsize$\pm$0.07} & 33.61{\scriptsize$\pm$0.46} & 37.97{\scriptsize$\pm$0.60} \\
\gr +MaCa  & 56.08{\scriptsize$\pm$0.15} & 46.28{\scriptsize$\pm$0.15} & 45.51{\scriptsize$\pm$0.62} & 
              51.95{\scriptsize$\pm$0.59} & 44.79{\scriptsize$\pm$0.75} & 39.26{\scriptsize$\pm$0.27} & 
              33.26{\scriptsize$\pm$0.15} & 33.58{\scriptsize$\pm$0.25} & 39.83{\scriptsize$\pm$0.37} \\ \midrule
GPTAQ & 55.39{\scriptsize$\pm$0.86} & 43.30{\scriptsize$\pm$0.47} & 34.15{\scriptsize$\pm$0.31} & 
          54.50{\scriptsize$\pm$0.98} & 44.31{\scriptsize$\pm$0.95} & 38.91{\scriptsize$\pm$0.09} & 
          43.04{\scriptsize$\pm$0.41} & 36.43{\scriptsize$\pm$0.41} & 38.59{\scriptsize$\pm$0.52} \\
\gr +MaCa  & 57.21{\scriptsize$\pm$0.30} & 51.24{\scriptsize$\pm$0.11} & 39.70{\scriptsize$\pm$0.93} & 
              54.97{\scriptsize$\pm$1.28} & 46.20{\scriptsize$\pm$2.70} & 38.94{\scriptsize$\pm$0.21} & 
              45.75{\scriptsize$\pm$0.56} & 39.22{\scriptsize$\pm$0.96} & 39.04{\scriptsize$\pm$1.45} \\ \midrule  \midrule
\textbf{Method} & \multicolumn{3}{c|}{\textbf{Qwen3-8B-IT}} & \multicolumn{3}{c|}{\textbf{Gemma3-12B-IT}} & \multicolumn{3}{c}{\textbf{LLaMA3.1-8B-IT}} \\ \midrule
& 4bit & 3bit & 2bit/g128 & 4bit & 3bit & 2bit/g128 & 4bit & 3bit & 2bit/g128 \\ \midrule
GPTQ   & 58.40{\scriptsize$\pm$0.26} & 43.96{\scriptsize$\pm$0.98} & 40.62{\scriptsize$\pm$0.09} & 
          57.08{\scriptsize$\pm$1.60} & 47.46{\scriptsize$\pm$3.10} & 38.33{\scriptsize$\pm$0.91} & 
          35.16{\scriptsize$\pm$0.14} & 34.65{\scriptsize$\pm$0.33} & 41.28{\scriptsize$\pm$0.19} \\
\gr +MaCa  & 60.96{\scriptsize$\pm$0.34} & 49.80{\scriptsize$\pm$1.30} & 49.39{\scriptsize$\pm$1.89} & 
              58.84{\scriptsize$\pm$0.53} & 51.21{\scriptsize$\pm$1.06} & 38.64{\scriptsize$\pm$0.48} & 
              35.31{\scriptsize$\pm$0.42} & 35.80{\scriptsize$\pm$0.93} & 44.52{\scriptsize$\pm$1.04} \\ \midrule
GPTAQ & 60.92{\scriptsize$\pm$0.32} & 47.21{\scriptsize$\pm$1.45} & 43.02{\scriptsize$\pm$1.00} & 
          59.88{\scriptsize$\pm$1.47} & 52.06{\scriptsize$\pm$2.95} & 38.96{\scriptsize$\pm$0.77} & 
          52.06{\scriptsize$\pm$0.31} & 41.97{\scriptsize$\pm$0.17} & 41.67{\scriptsize$\pm$1.31} \\
\gr +MaCa  & 61.89{\scriptsize$\pm$0.41} & 55.88{\scriptsize$\pm$1.00} & 44.61{\scriptsize$\pm$1.38} & 
              61.43{\scriptsize$\pm$0.18} & 54.81{\scriptsize$\pm$0.56} & 40.78{\scriptsize$\pm$1.42} & 
              54.11{\scriptsize$\pm$0.83} & 46.57{\scriptsize$\pm$0.57} & 42.35{\scriptsize$\pm$0.62} \\
\bottomrule
\end{tabular}
}
\label{tab:main-result}
\end{table*}

To address, we propose MaCa (\underline{Ma}tryoshka \underline{Ca}libration), a simple yet effective method that redefines the aggregation of calibration statistics. MaCa incorporates multi-scale sequence length information into Hessian estimation and treats each sequence as an independent sample. This yields a more stable and representative Hessian for quantization. By combining short and long sequences with normalization, MaCa produces a richer Hessian, which improves GPTQ update step. To the best of our knowledge, we are the first to systematically address the impact of sequence length variability within the PTQ of LLMs.

To summarize, our main contributions are as follows.
\begin{itemize}[itemsep=0pt, topsep=0pt, leftmargin=1.5em]
    \item We propose MaCa(\ref{alg:pseudo}), a novel Hessian estimation method that accounts for sequence length variability, treating all samples of different lengths with equal importance.
    \item Through extensive experiments, MaCa consistently improves the performance of Hessian-based quantization (e.g., GPTQ, GPTAQ) across a wide range of models, with clear improvements in various downstream tasks. (Table~\ref{tab:main-result}, Table~\ref{tab:longbench}, Table~\ref{tab:ablation})
    \item We show that MaCa constructs a richer Hessian by integrating short and long sequences, leading to more effective weight updates and lower quantization error. (Figure~\ref{fig:main-figure}, Figure~\ref{fig:recon-error})
\end{itemize}
\vspace{-0.5em}

\section{Background}

Hessian-based PTQ relies on a second-order Taylor expansion of the loss to minimize the perturbations introduced by quantization. Specifically, for a small weight perturbation $\Delta w$, the loss difference can be approximated as
\[
\Delta \mathcal{L} \approx \tfrac{1}{2}\Delta w^\top H_{(w)}\Delta w,
\]
where $H_{(w)} = \mathbb{E}[\nabla_w^2 \mathcal{L}]$ denotes the Hessian of the loss with respect to the weights. Based on this approximation, Nagel et al.~\cite{nagel2020adaround} established the foundation for Hessian-based PTQ methods by decomposing the Hessian $H_{(w)}$ of a linear layer ($z = Wx$) into input and output components, $H_{\mathrm{in}} \otimes H_{\mathrm{out}}$.

\subsection{Foundational Hessian-Based PTQ: GPTQ}

GPTQ \cite{frantar2023gptq} dramatically simplifies the computation of low bit quantization of LLMs by simplifying Hessian. Specifically, it assumes that the output side Hessian, $H_{out}$, is an identity matrix and further reduces costs by employing a column-wise quantization scheme with a Cholesky update. This backpropagation free approach provided a practical alternative to earlier gradient-based methods (e.g., \cite{nagel2020adaround}) that were too time-consuming for LLMs.

Subsequent research has mainly focused on improving the accuracy of GPTQ by developing better estimates for output side Hessian, $H_{out}$. BoA \cite{kim2025boa} incorporated inter-layer dependencies within the attention module by using an attention reconstruction error for $H_{out}$. In a slightly different direction, GPTAQ \cite{li2025gptaq} further improves GPTQ by introducing asymmetric calibration, which matches each quantized layer’s output to the full precision output, thereby reducing error accumulation across layers.

\subsection{The Overlooked Impact of $H_{in}$}
We introduce a fundamentally different approach, shifting the focus from improving $H_{out}$ to $H_{in}$. As visualized in Figure~\ref{fig:main-figure}, the activation distributions of LLMs are highly dependent on the input sequence length. Specifically, the length of the input sequence has a profound impact on $H_{in}$ that has not been a main focus so far. 

For an input sequence of length $L$, we denote hidden states as $x_t \in \mathbb{R}^D$ and collect them into $X \in \mathbb{R}^{D \times L}$. The input-side Hessian is then $H_{in}(L) := \mathbb{E}[X X^\top \mid L]$. Conventional PTQ methods adopt a single calibration length $L_C$, 
which restricts the Hessian to a narrow slice of information and yields a limited representation of model behavior across diverse deployment contexts.

\begin{algorithm}[t]
\caption{Hessian Calculation with MaCa}
\label{alg:pseudo}
\begin{algorithmic}[1]
\Require Calibration dataset $\mathcal{D}_{\text{calib}} = \{X_1, \dots, X_M\}$ of input activations for a layer $\mathcal{F}$, where each $X_i \in \mathbb{R}^{D \times L_m}$ has a variable sequence length $L_m$.
\State $D \gets \text{input dimension of } \mathcal{F}$
\State $H \gets \mathbf{0} \in \mathbb{R}^{D \times D}$ \Comment{Initialize Hessian}
\For{$i = 1, \dots, M$}
    \State $X_i \gets \text{Get } i\text{-th activation matrix from } \mathcal{D}_{\text{calib}}$
    \State $H_m \gets \frac{1}{L_m} (X_m X_m^\top)$ \Comment{Length-normalized}
    \State $H \gets \frac{m-1}{m} H + \frac{1}{m} H_m$ 
    \State $m \gets m + 1$    
\EndFor
\State \Return $H$
\end{algorithmic}
\end{algorithm}

\section{Method: Matryoshka Calibration}
\label{sec:method}
We now introduce MaCa, a simple yet effective method that produces fruitful Hessian in terms of various sequence lengths. The key insight is to change how calibration statistics are aggregated. In addition, instead of token level averaging, which gives longer sequences disproportionate influence, we treat each calibration sample as equally important. This produces a richer covariance that reflects the weight's input-channel quantization sensitivity.

\subsection{Multi-Length Aggregation}

In GPTQ, the Hessian is typically estimated using calibration data of a fixed length (e.g., 2048 tokens). MaCa adopts a multi-length calibration strategy to better reflect the  variable sequence lengths encountered during inference. Specifically, for each calibration batch, lengths are drawn uniformly from a predefined set (e.g., ${[256, 512, 1024, 2048, 4096]}$). This simple modification ensures that the aggregated Hessian incorporates information across diverse sequence lengths, alleviating the bias inherent in fixed length calibration.

\subsection{Length Agnostic Normalization}

GPTQ allows multi-length aggregation, but its token weighted Hessian update inherently overweights longer sequences, biasing $H$. Specifically, for a batch of input activations $X \in \mathbb{R}^{D \times L}$ (with $L$ tokens), the Hessian $H$ is updated via a moving average in GPTQ:
\begin{equation}
    H_{i} = \beta \cdot H_{i-1} + \alpha \cdot (X X^\top)
\end{equation}
where the weights $\beta = \frac{N_{\text{old}}}{N_{\text{old}} + L}$ and $\alpha = \frac{L}{N_{\text{old}} + L}$ depend on the total number of tokens processed so far, $N_{\text{old}}$. This formulation inherently biases the estimate. A calibration sample with more tokens contributes more to the Hessian than one with fewer tokens, thereby skewing it toward the statistical properties of longer sequences.

To overcome this, MaCa replaces the token level moving average with a sample level formulation. For a calibration set of $M$ sequences ${X_1, X_2, \dots, X_M}$, we update the Hessian as
\begin{equation}
    H_{i} = \left(\frac{m-1}{m}\right) H_{i-1} + \left(\frac{1}{m}\right) \left(\frac{1}{L_m} X_m X_m^\top\right)
\end{equation}
Here, the update weights are determined by the sample count $m$, not the token count. Each sample's contribution, $X_m X_m^\top$, is first normalized by its own length $L_m$, ensuring that every sequence, regardless of its length, has an equal impact on the final Hessian. This principled approach yields a more robust covariance estimate that better reflects the varied contexts a model will encounter post deployment. The procedure is summarized in Algorithm~\ref{alg:pseudo}.

\begin{table}[t]
\caption{\textbf{LongBench} overall scores (higher is better) for GPTQ and GPTAQ with and without MaCa on 4bit quantized models. Results are reported as mean ± std over 3 seeds.}
\small
\centering
\resizebox{0.48\textwidth}{!}{%
\begin{tabular}{l|c|c|c}
\toprule
\textbf{Method} & \textbf{Qwen3-4B-IT} & \textbf{Gemma3-4B-IT} & \textbf{LLaMA3.2-3B-IT} \\ \midrule
GPTQ   & 6.07{\scriptsize$\pm$1.27} & 8.29{\scriptsize$\pm$0.52} & 0.14{\scriptsize$\pm$0.06} \\
\gr + MaCa  & 8.31{\scriptsize$\pm$1.06} {\color{BrickGreen}(+2.24)} & 9.49{\scriptsize$\pm$0.36} {\color{BrickGreen}(+1.20)} & 0.14{\scriptsize$\pm$0.13} {\color{BrickGray}(+0.00)} \\ \midrule
GPTAQ & 10.81{\scriptsize$\pm$0.22} & 9.71{\scriptsize$\pm$0.32} & 3.64{\scriptsize$\pm$0.02} \\
\gr + MaCa  & 11.12{\scriptsize$\pm$1.62} {\color{BrickGreen}(+0.31)} & 9.72{\scriptsize$\pm$0.47} {\color{BrickGreen}(+0.01)} & 3.84{\scriptsize$\pm$0.18} {\color{BrickGreen}(+0.20)} \\
\bottomrule
\end{tabular}
}
\label{tab:longbench}
\end{table}

\section{Experiments}
\subsection{Experimental Setup}
\noindent \textbf{Setup.} All our experiments have been conducted with a single NVIDIA H100 GPU (80~GB). 

\noindent \textbf{Models.} We evaluate our method on three State-of-the-art LLM families: LLaMA3 \cite{grattafiori2024llama}, Qwen3 \cite{yang2025qwen3}, and Gemma3 \cite{team2025gemma}. For each LLM family, we consider both on-device (4B) and server-scale sizes (8B, 12B).

\noindent \textbf{Datasets.} We use the C4 training split for calibration. For benchmark, we use the average accuracy across eight zero-shot tasks: BoolQ \cite{clark2019boolq}, PIQA \cite{bisk2020piqa}, SIQA \cite{sap2019socialiqa}, HellaSwag \cite{zellers2019hellaswag}, WinoGrande \cite{sakaguchi2020winogrande}, ARC-easy \& challenge \cite{clark2018arc}, OpenBookQA \cite{mihaylov2018openbookqa}. For the LongBench, we use MultiFieldQA-en/zh \cite{bai2024multifieldqa}, NarrativeQA \cite{kocisky2018narrativeqa}, Qasper \cite{dasigi2021qasper}.

\noindent \textbf{Base Algorithms.} We apply MaCa to two hessian based quantization methods: GPTQ \cite{frantar2023gptq} and GPTAQ \cite{li2025gptaq}. We consider three precision settings: 4bit, 3bit, and 2bit. For 2bit quantization, we adopt a group size of 128, where weights are quantized in groups of 128 input channels that share the same quantization parameters. This group-wise strategy is commonly used to stabilize extremely low bit quantization. 

\noindent \textbf{Quantization Configuration.} We mostly follow GPTQ’s setup. Specifically, GPTQ estimates the Hessian using calibration sequences of fixed length 2048 with 256 samples. For a fair comparison, we fix the total number of calibration tokens to 524,288 (equivalent to 256 sequences of length 2048) for both base algorithms and MaCa. We use symmetric quantization and determine quantization parameters by minimizing $\| \Delta W \|_{F}^{2}$, a common approach in quantization fields. This objective follows the foundational premise of Hessian-based PTQ \cite{nagel2020adaround, frantar2023gptq}, where accurate second-order information is essential for minimizing output perturbation.


\begin{table}[t]
\caption{\textbf{Ablation} of {MaCa}’s multi-scale length aggregation (“+Multi-Scale”) and per-sequence normalization (“+Normalization”) sequentially starting from GPTQ and GPTAQ at 4bit, 3bit, and 2bit/g128.}

\small
\centering
\resizebox{0.48\textwidth}{!}{%
\begin{tabular}{l|c|c|c}
\toprule
\textbf{Method} & \textbf{4bit} & \textbf{3bit} & \textbf{2bit/g128} \\ \midrule
GPTQ   & 54.90 & 41.73 & 42.35 \\
+Multi-Scale & 56.00 {\color{BrickGreen}(+1.10\%p)} & 42.65 {\color{BrickGreen}(+0.92\%p)} & 44.81 {\color{BrickGreen}(+2.46\%p)} \\
+Normalization & 56.25 {\color{BrickGreen}(+1.35\%p)} & 46.45 {\color{BrickGreen}(+4.72\%p)} & 45.03 {\color{BrickGreen}(+2.68\%p)} \\ \midrule
GPTAQ & 55.92 & 42.86 & 33.22 \\
+Multi-Scale & 56.62 {\color{BrickGreen}(+0.70\%p)} & 46.99 {\color{BrickGreen}(+4.13\%p)} & 37.36 {\color{BrickGreen}(+4.14\%p)} \\
+Normalization & 56.93 {\color{BrickGreen}(+1.01\%p)} & 51.21 {\color{BrickGreen}(+8.35\%p)} & 39.15 {\color{BrickGreen}(+5.93\%p)} \\
\bottomrule
\end{tabular}
}
\vspace{-0.5em}
\label{tab:ablation}
\vspace{-0.5em}
\end{table}

\subsection{Main Results}
In Table~\ref{tab:main-result}, we show that {MaCa} consistently improves the average downstream tasks under the same calibration token budget across three LLM families (Qwen3, Gemma3, and LLaMA3). For instance, on Qwen3-8B we observe +5.5 points at 3bit and +8.7 points at 2bit/g128 applied on GPTQ. When layered on GPTAQ, {MaCa} further lifts accuracy and retains the same trend. Overall, {MaCa} closes a substantial portion of the performance gap introduced by low bit quantization while adding no extra calibration budget and requiring only a change in how statistics are aggregated with length agnostic normalization.

\subsection{Evaluation on LongBench}
We additionally evaluate our method on LongBench, a multi-task benchmark for long context understanding. Specifically, we focus on single document QA tasks including MultiFieldQA‑en/zh (average length 4.6k, 6.7k length), NarrativeQA (18.4k), and Qasper (3.6k), with performance measured with F1 score. As shown in Table~\ref{tab:longbench}, {MaCa} makes consistent improvements over both GPTQ and GPTAQ indicating MaCa's length-aware Hessian .

\subsection{Ablation Study}
In Table~\ref{tab:ablation}, we sequentially add our key algorithmic components ((1) multi-scale length aggregation (2) per-sequence normalization) to validate each effects. We observe that both components make nontrivial contributions for achieving higher performance. In particular, multi-scale aggregation alone improves GPTQ by up to +2.46\%p and GPTAQ by +4.14\%p, while adding per sequence normalization makes further gains across all settings. To demonstrate generality, we also applied MaCa to the activation-aware AWQ \cite{lin2024awq}. On Qwen3-4B (4bit), MaCa improved average zero-shot accuracy from 59.93\% to 60.28\%, confirming its effectiveness beyond Hessian-based frameworks.

\subsection{Reconstruction Error}
\begin{figure}[t]
    \centering
    \includegraphics[width=0.49\textwidth]{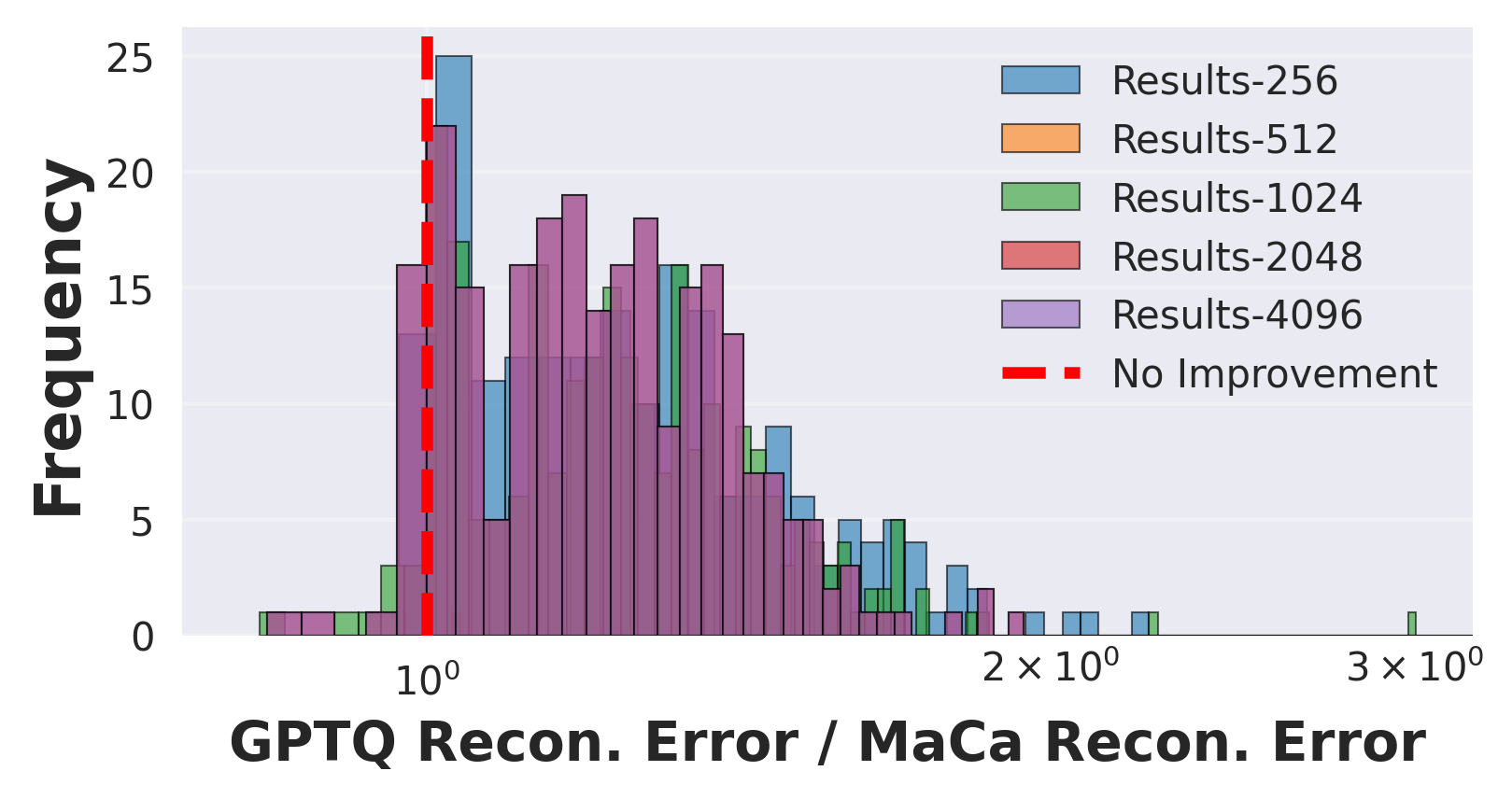}
    \vspace{-1.2em}
    \caption{\textbf{Ratio of Reconstruction Error (GPTQ / {MaCa})}. Histogram of error ratios across all linear layers of Qwen3-4B with 4bit quantized. Values $>1$ mean {MaCa} has lower reconstruction error, which leads to better quantization.}
    \label{fig:recon-error}
\vspace{-1em}
\end{figure}

In Figure~\ref{fig:recon-error}, we present a histogram of the ratio of GPTQ's reconstruction error to MaCa's reconstruction error across all layers with various sequence lengths of Qwen3-4B model quantized to 4bit. We measure the layer-wise reconstruction error, defined as the Frobenius norm of the quantization error, $\| \Delta WX \|_F^2$. This metric directly assesses how well the quantized model preserves the original model's outputs to provide a more fundamental evaluation of our method's effectiveness. The distribution is heavily skewed to the right of the ``No Improvement'' line, demonstrating that MaCa consistently achieves lower reconstruction error than GPTQ, which means that MaCa’s superior performance on  benchmarks is not incidental but a direct consequence of a fundamentally more accurate quantization.

\section{Conclusion}
In this work, we highlighted and quantified that Hessian in PTQ is fundamentally dependent on sequence length, a factor overlooked by existing works. To address, we introduced MaCa, a simple yet effective method that integrates multi-scale sequence lengths and treats each sample independently by normalization. Through extensive experiments on diverse benchmarks, we demonstrated that our method consistently improves GPTQ and GPTAQ, with clear gains in low bit settings and long context robustness. In contrast with prior approaches that refine the $H_{out}$, MaCa is the first to alter the $H_{in}$ accounting for length variability, producing richer Hessian without additional training or calibration cost. 


\bibliographystyle{IEEEbib}
\bibliography{strings,refs}

@STRING{NeurIPS = "Advances in Neural Information Processing Systems"}

@STRING{ICLR   = "International Conference on Learning Representations"}

@STRING{ICML   = "Proceedings of the International Conference on Machine Learning"}

@String {AAAI   = "Proceedings of the AAAI Conference on Artificial Intelligence"}

@STRING{ACL    = "Proceedings of the Annual Meeting of the Association for Computational Linguistics"}

@STRING{EMNLP  = "Conference on Empirical Methods in Natural Language Processing"}

@STRING{NAACL  = "NAACL"}

@inproceedings{son2024prefixing,
  author    = {S.~Son and W.~Park and W.~Han and K.~Kim and J.~Lee},
  title     = {Prefixing Attention Sinks can Mitigate Activation Outliers for Large Language Model Quantization},
  booktitle = EMNLP,
  year      = {2024},
}

@inproceedings{kim2025boa,
  author    = {J.~Kim and H.~Kim and E.~Cho and C.~Lee and J.~Kim and Y.~Jeon},
  title     = {{BOA}: Attention-aware Post-training Quantization without Backpropagation},
  booktitle = ICML,
  year      = {2025}
}

@inproceedings{kim2024towards,
  author    = {J.~Kim and C.~Lee and E.~Cho and K.~Park and H.~Kim and J.~Kim and Y.~Jeon},
  title     = {Towards Next-Level Post-Training Quantization of Hyper-Scale Transformers},
  booktitle = NeurIPS,
  year      = {2024},
}

@inproceedings{frantar2023gptq,
  author    = {E.~Frantar and S.~Ashkboos and T.~Hoefler and D.~Alistarh},
  title     = {{GPTQ}: Accurate Post-Training Quantization for Generative Pre-trained Transformers},
  booktitle = ICLR,
  year      = {2023}
}

@inproceedings{li2025gptaq,
  author    = {Y.~Li and R.~Yin and D.~Lee and S.~Xiao and P.~Panda},
  title     = {{GPTAQ}: Efficient Finetuning-Free Quantization for Asymmetric Calibration},
  booktitle = ICML,
  year      = {2025}
}

@inproceedings{nagel2020adaround,
  author    = {M.~Nagel and R.~A.~Amjad and M.~van Baalen and C.~Louizos and T.~Blankevoort},
  title     = {Up or Down? Adaptive Rounding for Post-Training Quantization},
  booktitle = ICML,
  year      = {2020}
}

@article{agarwal2025gpt,
  author    = {S.~Agarwal and L.~Ahmad and J.~Ai and S.~Altman and A.~Applebaum and E.~Arbus and R.~K.~Arora and Y.~Bai and B.~Baker and H.~Bao and others},
  title     = {gpt-oss-120b \& gpt-oss-20b Model Card},
  journal   = {arXiv preprint arXiv:2508.10925},
  year      = {2025}
}

@article{grattafiori2024llama,
  author    = {A.~Grattafiori and A.~Dubey and A.~Jauhri and A.~Pandey and A.~Kadian and A.~Al{-}Dahle and A.~Letman and A.~Mathur and A.~Schelten and A.~Vaughan and others},
  title     = {The llama 3 herd of models},
  journal   = {arXiv preprint arXiv:2407.21783},
  year      = {2024}
}

@article{yang2025qwen3,
  author    = {A.~Yang and A.~Li and B.~Yang and B.~Zhang and B.~Hui and B.~Zheng and B.~Yu and C.~Gao and C.~Huang and C.~Lv and others},
  title     = {Qwen3 technical report},
  journal   = {arXiv preprint arXiv:2505.09388},
  year      = {2025}
}

@article{team2025gemma,
  author    = {Gemma~Team and A.~Kamath and J.~Ferret and S.~Pathak and N.~Vieillard and R.~Merhej and S.~Perrin and T.~Matejovicova and A.~Ram{\'e} and M.~Rivi{\`e}re and others},
  title     = {Gemma 3 technical report},
  journal   = {arXiv preprint arXiv:2503.19786},
  year      = {2025}
}

@inproceedings{li2021brecq,
  author    = {Y.~Li and R.~Gong and X.~Tan and Y.~Yang and P.~Hu and Q.~Zhang and F.~Yu and W.~Wang and S.~Gu},
  title     = {{Brecq}: Pushing the Limit of Post-Training Quantization by Block Reconstruction},
  booktitle = ICLR,
  year      = {2021}
}

@inproceedings{clark2019boolq,
  author    = {C.~Clark and K.~Lee and M.-W.~Chang and T.~Kwiatkowski and M.~Collins and K.~Toutanova},
  title     = {BoolQ: Exploring the Surprising Difficulty of Natural Yes/No Questions},
  booktitle = NAACL,
  year      = {2019}
}

@inproceedings{bisk2020piqa,
  author    = {Y.~Bisk and R.~Zellers and R.~Le~Bras and J.~Gao and Y.~Choi},
  title     = {{PIQA}: Reasoning about Physical Commonsense in Natural Language},
  booktitle = AAAI,
  year      = {2020}
}

@inproceedings{sap2019socialiqa,
  author    = {M.~Sap and H.~Rashkin and D.~Chen and R.~Le~Bras and Y.~Choi},
  title     = {Social {IQa}: Commonsense Reasoning about Social Interactions},
  booktitle = EMNLP,
  year      = {2019}
}

@inproceedings{zellers2019hellaswag,
  author    = {R.~Zellers and A.~Holtzman and Y.~Bisk and A.~Farhadi and Y.~Choi},
  title     = {HellaSwag: Can a Machine Really Finish Your Sentence?},
  booktitle = ACL,
  year      = {2019}
}

@inproceedings{sakaguchi2020winogrande,
  author    = {K.~Sakaguchi and R.~Le~Bras and C.~Bhagavatula and Y.~Choi},
  title     = {WinoGrande: An Adversarial Winograd Schema Challenge at Scale},
  booktitle = AAAI,
  year      = {2020}
}

@article{clark2018arc,
  author    = {P.~Clark and I.~Cowhey and O.~Etzioni and T.~Khot and A.~Sabharwal and C.~Schoenick and O.~Tafjord},
  title     = {Think You Have Solved Question Answering? Try {ARC}, the {AI2} Reasoning Challenge},
  journal   = {arXiv preprint arXiv:1803.05457},
  year      = {2018}
}

@inproceedings{mihaylov2018openbookqa,
  author    = {T.~Mihaylov and P.~Clark and T.~Khot and A.~Sabharwal},
  title     = {Can a Suit of Armor Conduct Electricity? A New Dataset for Open Book Question Answering},
  booktitle = EMNLP,
  year      = {2018}
}

@inproceedings{bai2024multifieldqa,
  author    = {Y.~Bai and X.~Lv and J.~Zhang and H.~Lyu and J.~Tang and Z.~Huang and others},
  title     = {LongBench: A Bilingual, Multitask Benchmark for Long Context Understanding},
  booktitle = ACL,
  year      = {2024},
}

@article{kocisky2018narrativeqa,
  author  = {T.~Ko{\v{c}}isk{\'y} and J.~Schwarz and P.~Blunsom and C.~Dyer and K.~M.~Hermann and G.~Melis and E.~Grefenstette},
  title   = {The Narrative{QA} Reading Comprehension Challenge},
  journal = {Transactions of the Association for Computational Linguistics},
  year    = {2018}
}

@inproceedings{dasigi2021qasper,
  author    = {P.~Dasigi and K.~Lo and I.~Beltagy and A.~Cohan and N.~A.~Smith and M.~Gardner},
  title     = {A Dataset of Information-Seeking Questions and Answers Anchored in Research Papers},
  booktitle = NAACL,
  year      = {2021}
}

@inproceedings{lin2024awq,
  author    = {J.~Lin and J.~Tang and H.~Tang and S.~Yang and W.-M.~Chen and W.-C.~Wang and G.~Xiao and X.~Dang and C.~Gan and S.~Han},
  title     = {{AWQ}: Activation-aware Weight Quantization for On-device {LLM} Compression and Acceleration},
  booktitle = {MLSys},
  year      = {2024}
}

\end{document}